# Reroute Prediction Service

Ítalo Romani de Oliveira[1], Samet Ayhan, Michael Biglin, Pablo Costas

Boeing Research & Technology

Euclides C. Pinto Neto

IEEE Member

*Abstract* — The cost of delays was estimated as $ 33 billion only in 2019 for the US National Airspace System (NAS), a peak value following a growth trend in past years. Aiming to address this huge inefficiency, we designed and developed a novel Data Analytics (DA) and Machine Learning (ML) system, which aims at reducing delays by proactively supporting re-routing decisions.

Given a time interval up to a few days in the future, the system predicts if a reroute advisory for a certain Air Route Traffic Control Center (ARTCC) or for a certain advisory identifier will be issued, which may impact the pertinent routes. To deliver such predictions, the system uses historical reroute data, collected from the System Wide Information Management (SWIM) data services provided by the FAA, and weather data, provided by the US National Centers for Environmental Prediction (NCEP), part of the National Oceanic and Atmospheric Administration (NOAA). The data is huge in volume, and has many items streamed at high velocity, uncorrelated and noisy. The system continuously processes the incoming raw data and makes it available for the next step where an interim data store is created and adaptively maintained for efficient query processing. The resulting data is fed into an array of ML algorithms, which compete for higher accuracy. The best performing algorithm is used in the final prediction, generating the final results. Mean accuracy values higher than 90% were obtained in our experiments with this system.

Our algorithm divides the area of interest in units of aggregation and uses temporal series of the aggregate measures of weather forecast parameters in each geographical unit, in order to detect correlations with reroutes and where they will most likely occur. Aiming at practical application, the system is formed by a number of microservices, which are deployed in the cloud, making the system distributed, scalable and highly available. We implemented a web-based prototype client application that is integrated with Boeing's world-wide popular flight planning tool, ForeFlight®

Due to its unique features, this system provides increased situational awareness and operational insights to airline managers and Air Navigation Service Providers (ANSPs). With this system, airline managers can perform more accurate strategic and tactical operations planning and decision making, which result in time and fuel savings for airlines. The ANSPs are likewise able to plan and re-plan more accurately the flight routes, leading to more efficient utilization of airspace and airport capacities.

## I. Introduction

The FAA Office of Aviation Policy and Plans (APO) analysis of 2019 show that the cost of delayed flights rose by 9.3 percent, from $30.2 to $33.0 billion, an increase of $2.8 billion in just one year, and confirming a growth trend in the preceding years. Most of this increase was due to inefficiency in the National Airspace System (NAS), caused by inability to accurately predict future states of airspace and airports. A considerable part of this delay is caused by occurrences of severe weather, particularly convective weather and severe turbulence, which require alternative routes that usually take longer time to fly due to the avoidance of pertinent airspace. When such a situation arises, the Traffic Management Coordinators (TMCs) from FAA follow a process to create reroute plans, based on the National Severe Weather Playbook [1]. As explained in [2], Playbook routes are a set of standard routes that ATC can utilize to fit a particular set of circumstances, when the preferred routes are not available. These routes were created to allow for rapid implementation as needed, and there are 5 different types of routes:

- Airport-specific routes – designed to manage traffic to or from specific airports
- Airway Closure routes – designed to be used when a primary airway is closed (typically due to severe weather)
- East to West Transcon routes – designed to manage traffic from the eastern US (primarily the northeast) to the western US
- Regional Routes – designed to be used for traffic between specific regions of the US
- West to East Transcon routes – designed to manage traffic from the western US to the eastern US (particularly the northeast)

These routes are the basis for creating a Severe Weather Avoidance Plan (SWAP), which is a coordinated solution with inputs from one or more air traffic facilities in the NAS, each one having an elaborated local plan, according to its particular strategy, for managing a severe weather event. The Air Traffic Control System Command Center (ATCSCC) makes this coordination via a planning telcon each 2 hours. The results of a telcon are published in form of advisories, such as the one illustrated in Fig. 1, which affects flights between the DC metro airports and NY State. An advisory like this specifies a table prescribing reroutes on the right-hand side to the origin-destination pairs on the left-hand side. There may be more than one reroute for each origin-destination pair. The header of the advisory contains other information such as the advisory name, which is useful for reusing the contents of an advisory multiple times; specifications of the affected area and universe of applicability; validity times; probability of extension and other useful data.

Weather is not the only reason for issuing reroute advisories. In fact, reroutes may occur due to excess volume, active military airspace, or other reasons. However, this work focuses on the weather-related reroutes, which are the most frequently occurring ones. Furthermore, it is easier to obtain input data for weather-related reroutes towards building a prediction capability.

---

[1] Corresponding author's email:
italo.romanideoliveira@boeing.com

```
ATCSCC ADVZY 003 DCC 12/08/11 ROUTE RQD
NAME: J109_WEVEL_MODIFIED
CONSTRAINED AREA: ZNY ZDC
REASON: WEATHER
INCLUDE TRAFFIC: BWI/DCA/IAD DEPARTURES TO
                 ALB/BDL/BOS/BUF/CYYZ/PVD/ROC/SYR/ZBW/ZEU
FACILITIES INCLUDED: CZY/ZBW/ZDC/ZEU/ZOB
FLIGHT STATUS: ALL_FLIGHTS
VALID: ETD 080020 TO 080300
PROBABILITY OF EXTENSION: LOW
REMARKS: USE THE ROUTES IF SWANN PALEO ARE IMPACTED
ASSOCIATED RESTRICTIONS: ZDC TO ZOB AOB FL220 20 MIT DESTINED TO
                         ZBW.  ZDC TO ZOB FOR ZEU AT ALTIIUDE, 20
                         MIT.
MODIFICATIONS:
ROUTES:

ORIG            DEST            ROUTE
----            ----            -----
BWI DCA IAD     SYR             >JERES J211 LEONI J109
                                WEVEL ELZ<
BWI DCA IAD     ALB             >JERES J211 LEONI J109
                                WEVEL ELZ ITH<
BWI DCA IAD     ZBW             >JERES J211 LEONI J109
                                WEVEL ELZ ITH ALB<
BWI DCA IAD     BOS             >JERES J211 LEONI J109
                                WEVEL ELZ ITH ALB GDM3<
BWI DCA IAD     PVD             >JERES J211 LEONI J109
                                WEVEL ELZ ITH ALB TEDDY3<
BWI IAD         ZEU             RERTE:JERES J211 LEONI J109
                                WEVEL ELZ ITH ALB THEN TO
                                INITIAL FIX FOR TRACK
BWI DCA IAD     BDL             >JERES J211 LEONI J109
                                WEVEL ELZ ITH RKA SWEDE1<

TMI ID: RRDCC003
080031-080300
11/12/08 00:31  DCCOPS./nfs/lxstn20
```

Fig. 1: FAA reroute advisory example **[2]**.

If a flight that was filed by an airline with the FAA is affected by a reroute advisory, and no other alternative was given by the airline, the TMC will assign one of the reroutes to that flight. However, it is also possible for an airline to act proactively and have its own reroute plan before an advisory is issued, by means of the Collaborative Trajectory Options Program (CTOP). With CTOP, the FAA delegates a great part of the reroute responsibility to the airlines [3]. This concept allows airline dispatchers to submit several alternative routes to the FAA by means of a Trajectory Option Set (TOS) when adverse weather conditions knock out their initially preferred route [4, 5]. If an airline can proactively detect where and when a reroute is likely to occur, and compute new routes in advance, this will greatly reduce the inefficiency of the reroute operations, so this work presents the development of the detection component of this capability.

This paper presents a solution for the problem of predicting reroute advisories in a future horizon of up to a few days, building upon the principles first published in [6]. An accurate prediction of the most likely reroutes to occur is a key capability for better decision making for the airlines participating in the CTOP. It is also imperative for airspace and airport managers to proactively improve decision making which yields fluidity and efficiency of the whole NAS.

Indeed, a demonstration of reroute prediction capability was achieved in [7], in which samples from a particular source of weather measurement data is used, namely the cloud echo top readings, covering the entire NAS. Scores of the model's predictive performance are presented as True Positive and True Negative rates, showing promising values for data comprising the three summer months 2011. This capability though is not fully predictive because it uses current weather measurement data in the same 1-hour window that the reroute advisory happens, thus the solution is restricted to this approximate time window. Another predictive capability that tackles the delay effect of reroute, but not the reroutes directly, is presented in [8], also using measurement data and not weather forecasts. The use of short-term forecast data of convective weather is in introduced in [9], with a two-hour time window, however this study does a descriptive correlation analysis and does not tackle prediction.

Aiming at a capability to predict a reroute advisory several hours before it happens, our solution uses data from the Global Forecasting System (GFS) [10] provided by the National Centers for Environmental Prediction (NCEP). The GFS model contains a multitude of weather parameters evolving in certain regular intervals of time, thus forming a 4-dimensional grid spanning up to 16 days in advance. The accuracy of this forecast is considered by the experts to be acceptable until the 3-day span, after which it decreases dramatically and becomes less useful. Given this situation, predictions in the 3-day time window would be more reliable, especially in the first 24 hours, but even with this reduced window it is possible to make decisions that increase the operational efficiency of an airline and propagate to the efficiency of the whole NAS.

Another driver for pursuing the reroute prediction capability is the realization of the Trajectory-Based Operations (TBO) concept [11], in which specifications of a flight's trajectory are provided in advance and updated often during the whole lifecycle of the flight. Each specification is issued at a certain moment in this lifecycle, with the levels of detail and accuracy compatible to that moment, reflecting the best knowledge available from that flight that will happen or is happening. This knowledge greatly improves predictability and strategic decision making in ATM, and will be enormously benefited by a reroute prediction capability that triggers early changes in the trajectories and contribute to more strategic decision-making.

This prediction capability is achieved by means of Machine Learning (ML) techniques. These and other ML techniques have been employed extensively for routing systems, including some of the references cited above, and another study that evaluates the acceptability by the human operators of reroutes provided by an automated rerouting tool [12] utilizing an optimization algorithm. Our solution does not predict an exact new route for a certain flight, but instead predicts the reuse of a named reroute advisory that was used by the TMC in a previous occasion, as that of Fig. 1, that may be issued with updated routes.

## II. DATA INPUTS FOR THE PREDICTION MODEL

Our reroute prediction service uses past reroute advisories provided by the FAA and weather data provided by the NCEP. For the reroute advisories, there is a web page that lists links to the advisories published in the same format as that of Fig. 1, however the collection of such data over time is not a reliable solution because it requires periodic refresh of the pages, using assumptions on page format and interactive objects, and performing differential checks on page contents. But there is a more reliable alternative, from the stand point of automatic collection, which uses the TFMData data streaming service from the FAA [13]. Traffic Flow Management (TFM) uses TFMData, which is a publish / subscribe data service with high volume/high speed output (it generates several Gigabytes per day), that integrates the System Wide Information Management (SWIM) architecture [14] supporting modern ATM, which has been used in many systems, we have developed to date [15, 16, 17]. The simple scheme for obtaining TFMData messages is illustrated in Fig. 2.

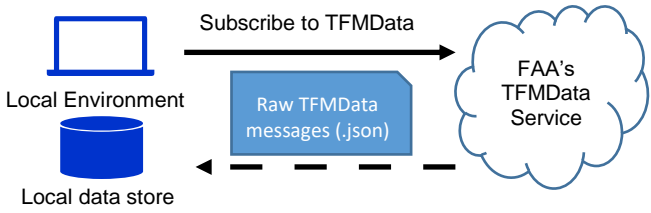

Fig. 2: Obtaining raw data from the TFMData service.

The reroute advisory messages are a subset of the TFMData messages called Traffic Management Initiatives (TMI) and need to be selected from the broadband stream. This data processing infrastructure eliminates unnecessary data and redundancies, creates indexes, establishes correlations and performs map-reduce capability, so that curated daily collections with content similar to that of TFMData is preserved while keeping it in database structures, which are optimized for efficient query processing. The raw data is deleted each day after it is processed.

The weather data is collected from the Global Forecasting System (GFS) [10]. This system periodically publishes GRIB files (General Regularly-distributed Information in Binary form), which are 4-dimensional grids in which, for each grid point, 600+ weather parameters are available. The spatial resolution of the output grid can be chosen from a minimum size of 15 nautical miles (0.25 degrees lat/lon), to a maximum size of 60 nautical miles (1.0 degrees lat/lon), which we chose. In the vertical dimension, it has 17 several vertical layers, and in the temporal dimension, a resolution of 3 hours along 16 days out. These GRIB files can be obtained via FTP or HTTP from a known and stable directory structure and are updated in regular intervals of 6 hours, thus can be collected periodically in a reliable manner by means of data robots, as illustrated in Fig. 3.

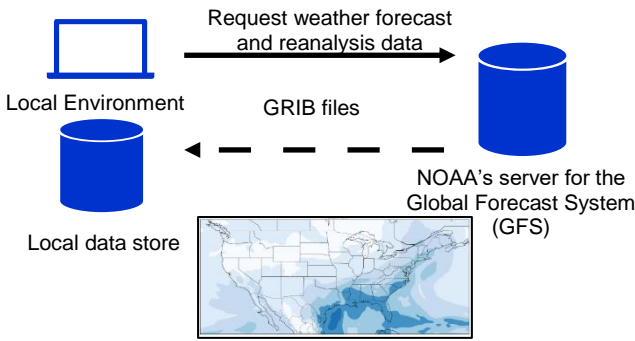

Fig. 3: Obtaining weather forecast data from NCEP.

There are other global forecast models available besides the NCEP's GFS. However, unlike NCEP's GFS these models may incur cost. In addition, these models are either optimized for specific regions of the world, or their GRIB files may be formatted in ways that cannot be readily manipulated by open-source libraries. Therefore NCEP's GFS remains our preferred choice for weather input data to our system.

## III. PIPELINE FOR MODEL TRAINING

The pipeline presented in Fig. 4 illustrates the set of steps from data collection to model evaluation. As described in the previous section, the Data Collection stage yields the "Local Dataset" data collections. Although the reroute messages pass through an intensive data processing, the weather data is still in a raw form and needs to be further processed to feed the prediction models. Hence, there is a second stage called "Data Wrangling", which is further expanded in Fig. 5.

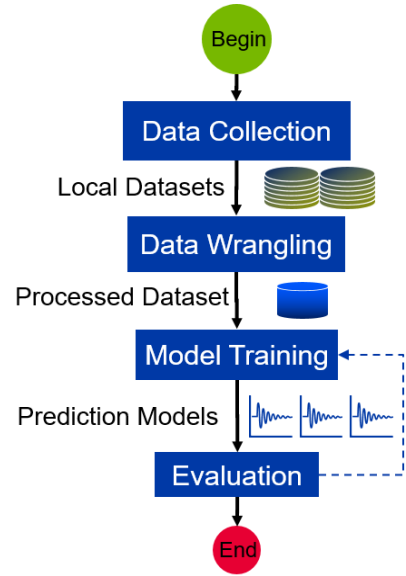

Fig. 4: pipeline for training the reroute prediction models.

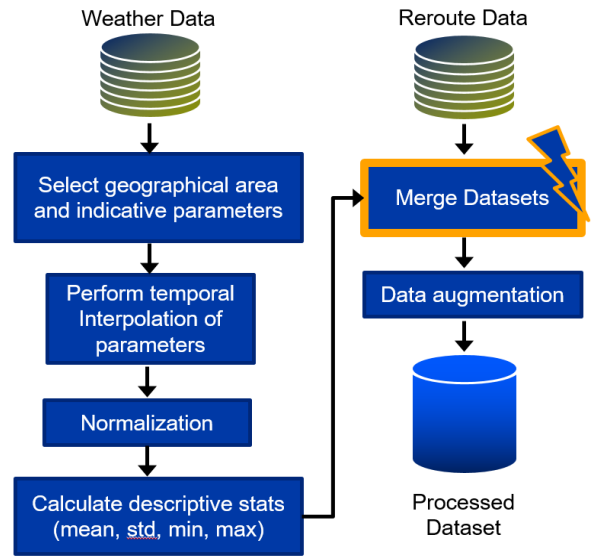

Fig. 5: Data wrangling pipeline.

In the data wrangling pipeline, the first step is to select a geographical area of interest from the global weather forecast model, otherwise we would be wastefully processing uncorrelated weather data. In that aspect, we developed two versions of the model, one per control area (ARTCC – Air Route Traffic Control Center), and another that covers the continental US (CONUS) as a whole. Thus, the explanation of the pipelines of Fig. 4 and Fig. 5 is presented below in duplicate sections, corresponding to each of the two system versions.

## IV. DATA WRANGLING PER ARTCC

The FAA reroute advisories indicate the ARTCCs affected by them in the field "CONSTRAINED AREA", as shown in Fig. 1, which indicates ZNY (New York area) and ZDC (DC area). The ARTCC is associated to a partition of the airspace as shown in Fig. 6 (2-dimensional top view). This partition induces a way of breaking down the prediction problem, because it sounds reasonable to assume that only the weather observed within an ARTCC will cause reroutes in that ARTCC. On the other hand, it can be argued that the surrounding weather is relevant too, however the exclusive use of intra-ARTCC weather immensely simplifies the ML models.

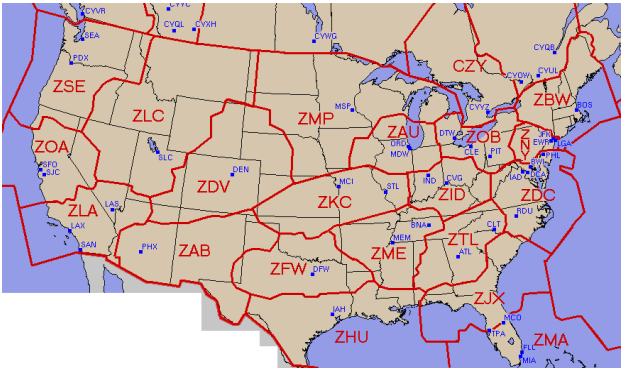

Fig. 6: ARTCC geographical division **[18]**.

Another simplification that greatly helps the model performance is to select only the most indicative weather parameters from the 600+ available in the GFS GRIB files. After many empirical trials, only the following ones turned out to be clearly correlated with reroute advisory occurrences: Precipitable Water (PW) [19, 20], Convective Available Potential Energy (CAPE) [21, 22], Air Potential Temperature (APT) [23, 24] and Convective Inhibition (CI) [25].

Furthermore, we assumed that the temporal evolution of the weather inside the ARTCC is indicative of reroute advisory issuances, so we augment the temporal resolution of the weather forecast data by interpolation. An interpolation is also necessary to better match the weather data with advisory validity times, hence there is the second step in the pipeline of Fig. 5. In this version, the interpolation resolution is of one minute. Then, a step of normalization is performed, where the long-term variation range of the weather parameters are used to transform the values to the interval between 0 and 1, for the sake of improving the performance of the ML models, which are designed to better work within that interval.

After normalization, a huge data reduction is performed by taking aggregate statistics of the several grid points inside the ARTCC. These are called "descriptive stats" in Fig. 5 and include mean, standard deviation, minimum, maximum and, in this first version, also 1$^{st}$ and 3$^{rd}$ quartiles and median. This achieves a reduction from many thousands of grid points inside an ARTCC to just a few variables of temporal series. Then, the resulting weather forecast data is matched with the reroute advisory times in the "Merge datasets" step of Fig. 5, and these tuples have the data in the right format for prediction. However, before they can actually be fed to the models, they undergo augmentation, according to the following explanation.

The reason for the "Data augmentation" step of Fig. 5 is that the data being processed is imbalanced for a ML problem of the classification type. Given the input weather parameters at a certain time interval, the model has to classify the occurrence (1) or not (0) of a reroute for an ARTCC, therefore each ARTCC uses its own separate model. Our training dataset provides labeled examples (0 or 1), thus we are talking about supervised learning. The major difficulty in our case is that the input data has much larger numbers of negative (0) than positive (1) examples for reroute advisory occurrences, hence being largely imbalanced. In our sample with two years of data, only 16% of the days turn out to have occurrence of a reroute advisory. This is illustrated in Fig. 7.

In this figure there is a comparison among the numbers of occurrences of the four most frequent reroute advisory names and the total number of days in two years. This graph is still very favorable to the reroutes, because an advisory is typically valid only for couple of hours in the day. So, if the dataset size were shown in number of hours or minutes, the rightmost bar would be so disproportionate, that we could barely visualize the size differences among the smaller bars.

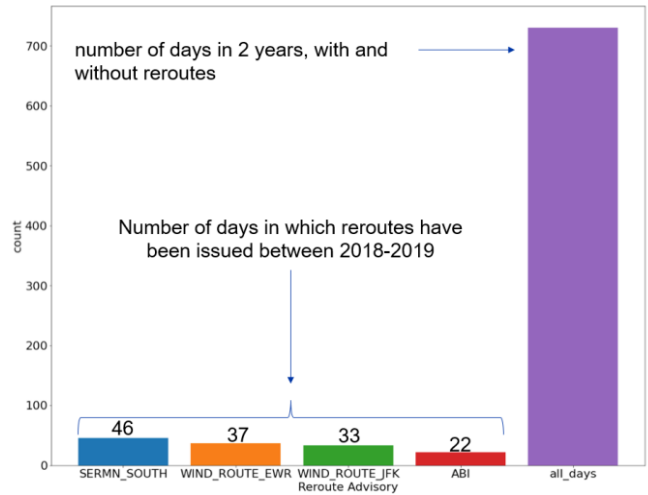

Fig. 7: Imbalance between occurrences and non-occurrences of reroutes.

Imbalanced datasets cause severe limitations in the accuracy of a ML predictor, thus we employed a combination of the techniques SMOTE and Tomek, referred to as SMOTETomek [26]. The SMOTE term stands for Synthetic Minority Over-sampling Technique [27] and creates additional positive observations of an output variable (a reroute advisory occurrence) by picking a real positive observation and using it in a convex combination of that observation with its nearest positive neighbors, for a given reroute advisory name. In our case, the convex combination is performed for the input weather parameters. Because these observations do not occur naturally in the data, they are called "synthetic", and make that the decision region of the positive examples be more general and less prone to over-fitting than simply oversampling the positive observations as they are. The selection of the nearest neighbors among the positive examples is performed by the so-called $k$-Nearest Neighbors ($k$-NN) algorithm [28], with the default value of $k$ given by the `imblearn` library [29], whose default is 5. After generating the synthetic data, the technique called Tomek links [30] is used. To explain it succinctly, a Tomek link is a pair of observations in which one of the members is in the majority class (output 0) and one in the minority class (1) and they are the nearest neighbors around. This proximity can confuse the ML algorithm and decrease its performance; thus such pairs of observations are eliminated from the data and we end up with more distinguishable separation boundaries between the output classes.

Having finished the data wrangling pipeline, the processed dataset is ready for model building, training and evaluation.

## V. MODEL BUILDING, TRAINING AND EVALUATION PER ARTCC

Building the models is a task performed in the development phase of the system, hence this task is not explicit in Fig. 4, which depicts the repetitive steps of the system once it is developed. Regardless of that, we must describe how the model is built, so we do it here, and we start with the very high level view of how the prediction models of this system work, as shown in Fig. 8.

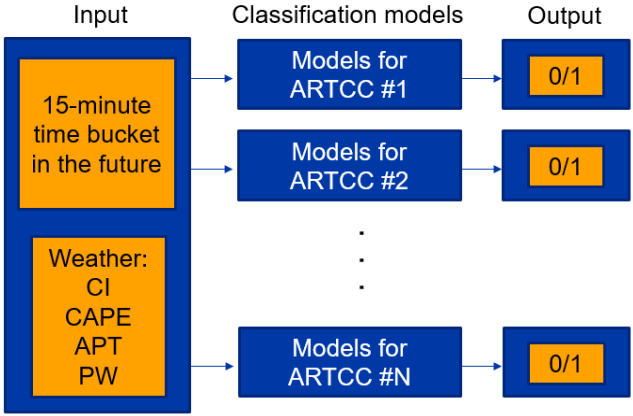

Fig. 8: High-level view of the prediction models per ARTCC.

Given a 15-minute time bucket (an index of a 15-minute interval) in the available weather forecast range, and the weather forecast for that time bucket (which with GFS can be given up to 16 days in advance), the system will run a classification model for each reroute advisory name and issue a prediction on whether an advisory will occur (1) or not (0). The choice of 15 minutes is due to this being a standard planning interval for the ATFM decision-making process. The word "models" is used in the plural because we run a number of different ML algorithms and select the best one for the output, in each case.

In this version of the system, the following ML algorithms were tested: Multilayer Perceptron, KNN, Bagging Classifier, Gradient Boosting, Extra-trees and Random Forest. We run these models in parallel and evaluate their performance with a scoring measure, explained below. The performance evaluation uses 20% of the input dataset as test data, separated from the training data, but in a cross-validation scheme, as illustrated in Fig. 9. This scheme alternates the selection of train and test data $k$ times (in this case, $k = 5$), in order to mitigate bias that a particular split can cause.

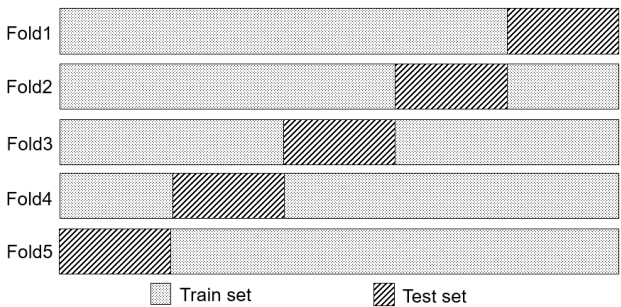

Fig. 9: Cross-validation of the prediction scores.

After this evaluation and tuning of the algorithm's configuration parameters, it is safe to re-train the model with the entire dataset, therefore there is an iterative tuning and selection process which is represented by the return arrow in Fig. 4.

The scoring measure for the algorithm is tailored, in order to better account for certain aspects of this problem. We named it as "reroute detection score", and it has this simple formulation:

Reroute Detection Score := (Eq. 1)
(Accuracy + Reroute Coverage) / 2

The Reroute Coverage term is the rate of reroute occurrences that were covered, independently of small differences in start and end time of the prediction. We deemed this beneficial because those small differences were driving down the accuracy measure and are of little relevance to the usefulness of the prediction capability. This can be illustrated by means of Fig. 10.

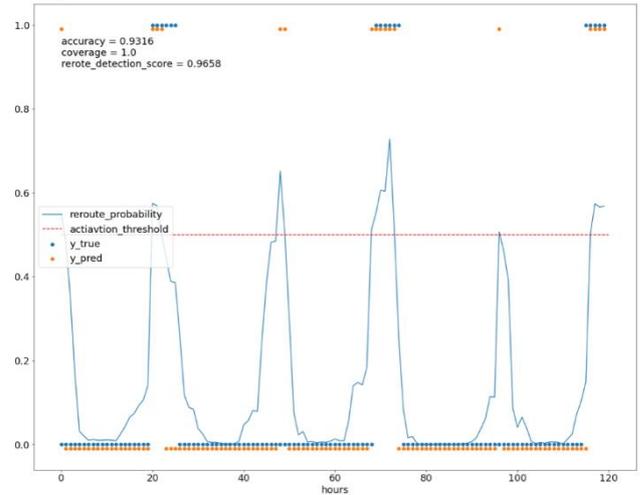

Fig. 10: Example of detection output and Reroute Detection Score.

In this figure, the spiked plot shows the output of the Random Forest prediction algorithm, which consists of a probability of class 1 or positive reroute occurrence. This first output passes through a simple threshold filter to signal whether a reroute will occur or not. In our case this filter is 0.5. The final prediction output `y_pred` thus is a binary variable outputted minute by minute, shown in orange dots in the graph, while the actual reroute occurrences `y_true` are shown in blue dots. The accuracy and reroute detection score are computed with these dot values (0 or 1). The coverage measure in this case is 1.0 because, if we consider each contiguous groups of `y_true=1` as a single occurrence of reroute advisory, all of them have some overlap with at least one occurrence of `y_pred=1`, that is, the algorithm is detecting all true positive events of reroute advisory.

Using this training and evaluation scheme for the dataset composed by the years 2018 and 2018, it turned out that the Random Forest algorithm systematically outperformed the other algorithms, with the following configuration parameters: 1000 estimators; `max_features=´log2´`; `criterion=´entropy´`; `random_state=64`; probability threshold = 0.5. The resulting reroute detection score per ARTCC is shown in Fig. 11.

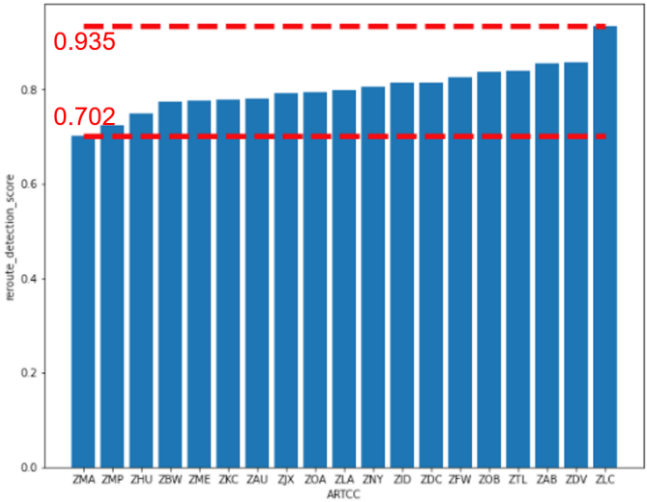

Fig. 11: Reroute Detection Score per ARTCC.

As it can be seen, the maximum score was obtained for the ZLC (Salt Lake City) ARTCC, and the minimum for ZMA (Miami). We so far have not investigated the reasons for the lower score cases.

## VI. Data Wrangling per Advisory Name

At some point in this research, we noted that the reroute advisories had a strongly repeating pattern and several of them involved more than one ARTCC. This is probably caused by the reliance of the TMCs on the National Severe Weather Playbook [1], already mentioned above. In fact, some reroute advisory names were reused many times and with little modification in the advisory contents. Thus, we developed a version of the prediction system centered on the reroute advisory name, instead of the ARTCC.

We refer again to the pipeline of Fig. 5 to describe the data processing aspects of this version. With respect to the selection of the geographical area, we start by selecting a rectangle covering the main contiguous US (CONUS) from the GRIB files, as shown in Fig. 12. Having that area selected, we select the same key weather parameters CI, CAPE, APT and PW as described in Section IV.

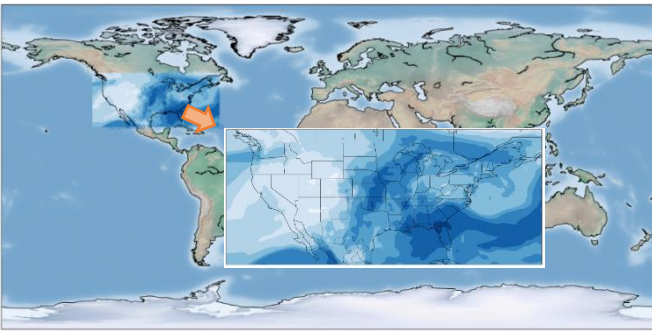

Fig. 12: Selection of the geographical area of interest in the second version.

The interpolation step here is done similarly to the previous version, however in steps of 15 minutes. Another difference here, but more crucial, is that, in the next step of the pipeline, which calculates descriptive statistics, the data aggregation is not performed per ARTCC nor for the whole coverage area, but with some intermediate aggregation. The GFS GRIB files provide values in a parallel grid with certain options of horizontal resolution, among which we chose the 60 nautical miles (1.0 degree lat/lon), in several vertical layers and a temporal resolution of 3 hours, along 16 days out. For a single 3-hour period and a single altitude layer in this grid, covering the CONUS requires thousands of grid points. Considering that the CONUS is contained in a rectangle of approximately 2328 nautical miles (nmi) of East-West length and 1374 nmi of North-South length, this includes approximately 890 grid cells and approximately the same number of grid points, and that is multiplied by 17 pressure altitude levels and 4 parameters, resulting in a figure of 60,473 input variables. But still there are several distinct reroute advisory names, with each one in fact requiring its own prediction model. Thus, to decrease the training time, we build a much coarser grid and, for each cell of this grid, calculate the descriptive statistics, as shown in Fig. 13, which shows the color scale plot of PW in the background. This aggregation reduces the size of the input dataset to 64 input variables for each detection time step.

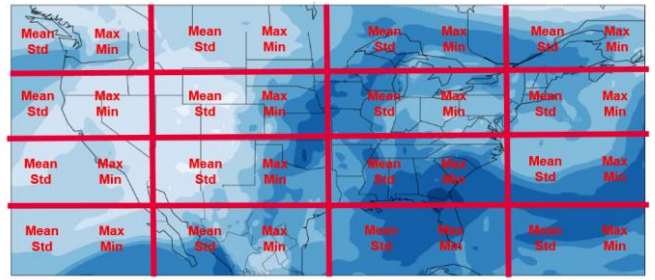

Fig. 13: Aggregation of weather data with statistics from large grid cells.

Having done this aggregation, the remaining steps of the data wrangling pipeline of Fig. 5 are done similarly to the previous version. However, the model will have a different meaning for the outputs, and different performance results, as shown in the next section.

## VII. Model Building, Training and Evaluation per Advisory Name

Here the classification models are oriented to predict a specific reroute advisory name, differently from the first version, which predicted any advisory name for a certain ARTCC. The overview of this new definition is provided in Fig. 14.

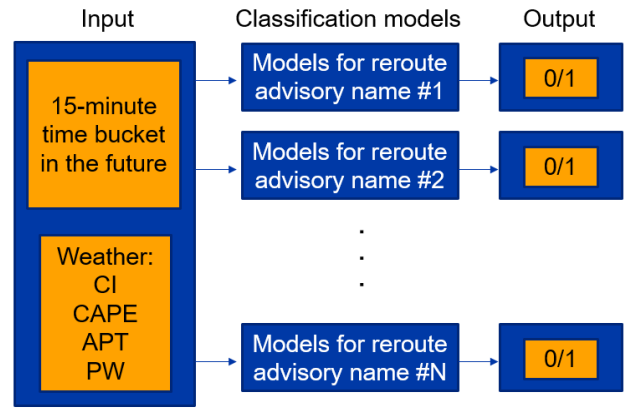

Fig. 14: High-level view of the prediction models per advisory name.

The following ML algorithms are used in this version: Multilayer Perceptron, Support Vector Machine (SVM), Gradient Boosting, and the Extra-trees Classifier.

The following model configurations were used:

- **Multilayer perceptron**: 1 hidden layer, 100 nodes, Rectified Linear (Relu) as activation function, 10% of validation fraction, Adam as optimizer, 0.001 as learning rate;

- **SVM**: `C=1.0`, `kernel` = Radial Basis Function (RBF);

- **Extra-trees Classifier**: `N_estimators=1000`, Gini Impurity `'gini'`) as the quality measurement function;

- **Gradient Boosting**: `N_estimators=100`, `learning_rate=0.1`, `num_leaves=31`.

By selecting the best algorithm in 5-fold cross-validation sessions, using the reroute detection score as main criterion, the performance results of Fig. 15 were obtained, taking the average of the four most frequent reroute advisory names (already shown in Fig. 4).

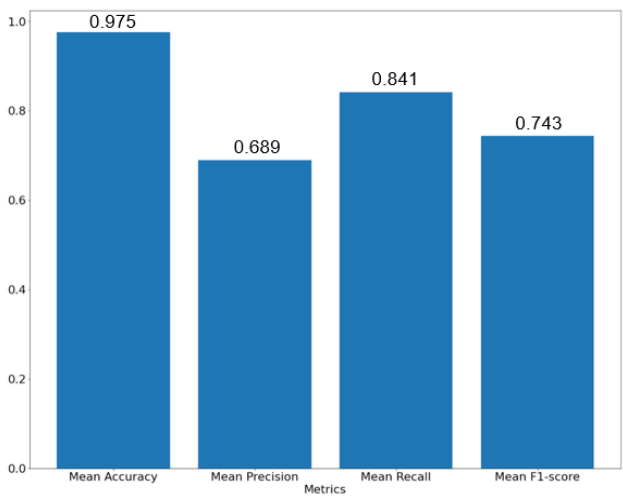

Fig. 15: Average performance metrics of the prediction models.

In this version of the prediction formulation, we noticed that the Extra-trees Classifier algorithm [31] tends to have the best performance. We used the freely available version of Extra-trees from the Scikit-learn module [32].

## VIII. Considerations for Commercial Implementation

We developed prototype implementations of the system that facilitate its deployment in the cloud as a set of microservices, aiming at higher flexibility, scalability and availability. These implementations can be invoked on any device with internet connection, including desktop and mobile devices. The service is exposed with a web-based client application that is integrated with Boeing's ForeFlight® app, a world-wide popular flight planning tool.

## IX. Final Remarks

Although having achieved satisfactory reroute detection scores, there are several directions of work that would help to increase the benefits offered by this system, and its overall applicability in practical settings:

- **Identification of which flight plans will be affected by a reroute and how they will be affected**: identification of the flight plans directly affected would be easy by using the route information in the reroute advisory. However, there may be indirect impacts due to congestion, and furthermore it would be desirable to automatically propose new flight plans that are compatible with the reroute advisories. A data-driven approach for these capabilities sounds the most promising way forward.

- **Integration to the CTOP process**: as a logical consequence of having the capability of predictively proposing new flight plans, these solutions could be leveraged by integration with CTOP.

- **Provision of dynamic confidence intervals for prediction scores**: it would be useful to output confidence intervals for the prediction scores, so that, when making a decision, the user could tune her/his reliance on the prediction model.

- **Using High-Performance Computing (HPC) to increase predictive power**: as shown in the paper, the present system does a lot of aggregations in the input variables in order to keep down the number of inputs and be manageable in a regular notebook computer. If we assume a HPC with powerful GPU units, large memory, and dozens of CPUs, the accuracy of the predictions in the ARTCCs with low accuracy might increase, and the prediction of rare reroute advisories might become acceptable. With the enhanced computing power, more years of data can be used in the training, less geographical aggregation can be employed, and that would supposedly lead to better performance in the difficult cases. The anticipative proposal of alternative routes would most certainly require HPC resources.

- **Extension to other regions of the world**: wherever reroute advisories can be collected, the same principles of this system can be applied. For example, the Eurocontrol Network Manager has the B2B rerouting service and is a clear possibility. But there may be other cases that we have not started to explore.